# TOPOLOGICAL STABILITY: A NEW ALGORITHM FOR SELECTING THE NEAREST NEIGHBORS IN NON-LINEAR DIMENSIONALITY REDUCTION TECHNIQUES

Mohammed Elhenawy[1], Mahmoud Masoud[1], Sebastian Glaser[1], Andry Rakotonirainy[1]

[1]Accident Research and Road Safety at the Queensland University of Technology, Australia,

*ABSTRACT*— In the machine learning field, dimensionality reduction is an important task. It mitigates the undesired properties of high-dimensional spaces to facilitate classification, compression, and visualization of high-dimensional data. During the last decade, researchers proposed many new (non-linear) techniques for dimensionality reduction. Most of these techniques are based on the intuition that data lies on or near a complex low-dimensional manifold that is embedded in the high-dimensional space. New techniques for dimensionality reduction aim at identifying and extracting the manifold from the high-dimensional space. Isomap is one of widely-used low-dimensional embedding methods, where geodesic distances on a weighted graph are incorporated with the classical scaling (metric multidimensional scaling). The Isomap chooses the nearest neighbours based on the distance only which causes bridges and topological instability. In this paper, we propose a new algorithm to choose the nearest neighbours to reduce the number of short-circuit errors and hence improves the topological stability. Because at any point on the manifold, that point and its nearest neighbours form a vector subspace and the orthogonal to that subspace is orthogonal to all vectors spans the vector subspace. The prposed algorithmuses the point itself and its two nearest neighbours to find the bases of the subspace and the orthogonal to that subspace which belongs to the orthogonal complementary subspace. The proposed algorithm then adds new points to the two nearest neighbours based on the distance and the angle between each new point and the orthogonal to the subspace. The superior performance of the new algorithm in choosing the nearest neighbours is confirmed through experimental work with several datasets.

## I. INTRODUCTION

Dimensionality reduction is the process of transforming a high-dimensional data into a meaningfully reduced dimensionality. The ideal goal of the dimension reduction is discovering the minimum number of parameters needed to account for the observed properties of the data which is called the intrinsic dimensionality of data [1]. Dimensionality reduction techniques take as input a collection of unorganized data points. The topology and intrinsic dimensionality of the underlying manifold are unknown and need to be estimated in the process of constructing a faithful low-dimensional embedding using the data sample. By representing the datasets using its reduced intrinsic dimension, many machine learning tasks such as classification, visualization, and compression of high-dimensional data become easier.

In the real-world, many datasets such as global climate patterns, stellar spectra, or human gene distributions, lie on complex nonlinear data manifolds and need nonlinear dimensionality reduction techniques that are capable of dealing with it. Joshua B. Tenenbaum et al proposed the Isomap algorithm which is a global approach for dimension reduction where nearby points in the high dimensional space should be nearby in the new reduced dimensional space and the same should be happen for the faraway points..Isomap is a nonlinear dimension reduction technique that is capable of discovering the nonlinear degrees of freedom that underlie complex natural observations, such as human handwriting or images of a face under different viewing conditions. The Isomap algorithm has three steps [2]. The first step is constructing weighted graph G over the data points by establishing the neighbourhood relationship between them. In the second step,Isomap estimates the geodesic distances between all pairs of points on the manifold $M$. These distances are the shortest path distances $d_G(i,j)$ in graph $G$ which are calculated by Floyd's algorithm. The Isomap's third step is finding the top $M$ eigenvectors of the matrix $(-HSH/2)$ where $S_{ij}$ is the square of the shortest distance between points $i$ and $j$ and $H$ is the centring matrix.

There are two types of Isomap; the first type is the unsupervised Isomap. Conformal Isomap (C-Isomap) is unsupervised Isomap which is developed to guarantee conformality [3]. Landmark Isomap (L-Isomap) is another unsupervised Isomap which is a computationally efficient approximation of Isomap developed to increase the speed of Isomap [4]. Incremental Isomap is a version of Isomap which is efficiently applied when data is collected sequentially[5]. The second type of Isomap is supervised Isomap. S-Isomap [6]and M-Isomap [7] are two supervised Isomap algorithms which utilize class information to guide the procedure of nonlinear dimensionality reduction. In all of the above versions of the Isomap, neighbourhood search is a common and an important step.

The simplest way to identifying the neighbourhood of each data point is identifying a fixed number of nearest neighbours, $k$, per data point, as measured by Euclidean distance or any other distance measure. The other criteria used to choose neighbours is based on the distance too, where all points that have distance smaller or equal to certain threshold are considered neighbours. For example, one can identify neighbours by choosing all points within a ball of fixed radius. The number of neighbours for each point could be different. The step of neighbourhood selection can be more sophisticated. As an example, all points within a certain radius up to some maximum number can be considered as neighbours. Another quite sophisticated method, we can take up to a certain number of neighbours but none outside a maximum radius. In general, specifying the neighbourhoods in Isomap depends only on distance information and does not incorporate any other information that may help to get better low dimensional embedding. Isomap estimating the nonlinear intrinsic geometry of a data manifold based on a rough estimate of each data point's neighbours on the manifold, hence the ability of Isomap to estimate the intrinsic geometry of a data manifold depends on choosing the correct neighbours for each point. Unfortunately, defining the connectivity of each data point via its nearest Euclidean neighbours in the high-dimensional is vulnerable to short-circuit errors [8]. These errors could happen if the folds in the manifold on which the data points lie are too close and the distances between these points are smaller than the true neighbourhood dstance. For example, a single data point which lies on one surface and very close to another surface could connect these two surfaces if the true neighborhood distance is relatively large . Many entries in the geodesic distance matrix can be altered by a single short-circuit error and consequently it leads to a drastically different (and incorrect) low-dimensional embedding. The left column in Figure (1) shows an example of how short-circuit errors lead to incorrect low dimensional embedding.On the other hand, the right column of Figure (1) shows no short-circuit errors and a correct low dimensional embedding. In this work, we incorporate another piece of information into the neighbourhood selection in order to ensure the topological stability. Since each point and its neighbour lie in low dimension vector subspace ($W$) we use the angle between the neighbouring candidate and the normal to the bases of that vector space ($W^\perp$). This angle should be close to 90 degrees to consider the candidate neighbour as neighbour.

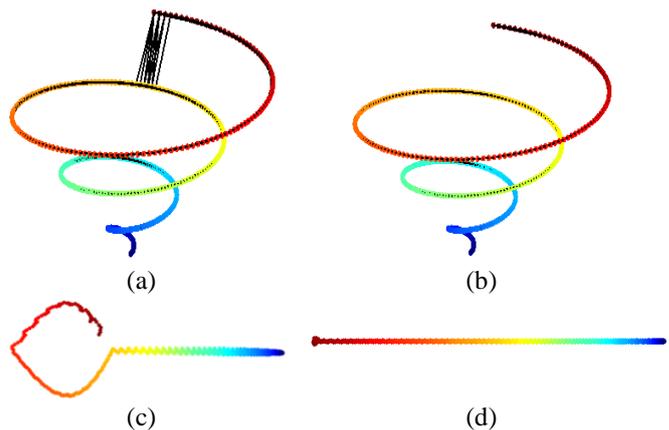

(a)  (b)

(c)  (d)

Figure (1) illustrates the topological instability because of the short-circuit errors that happened during a neighbourhood search step.

## II. THEORETICAL BACKGROUND AND METHODS [9,10]

Given a set of unorganized data points which are sampled from the vector space $V$ over the field R

$$V = R^D = \left\{ x = \begin{pmatrix} x_1 \\ \vdots \\ x_D \end{pmatrix} : x_j \in R \right\} \quad (1)$$

Let $W$ be a finite-dimensional subspace of a vector space $V$ which means the subspace $W$ has a finite basis. If $W$ is a finite-dimensional subspace of a vector space $V$ then all basis for $W$ will have the same number of vectors. The number of vectors in any basis is called the dimension of the subspace $W$. If $V$ itself has finite bases we say $V$ is a finite-dimensional vector space then the number of vectors in any basis is called the dimension of $V$.

Let $W^\perp$ denote the collection of all vectors in $R^D$ which are orthogonal to all the vectors in $W$

$$W^\perp = \{ x \in R^D : \langle x, w \rangle = 0 \ \forall w \in W \} \quad (2)$$

$W^\perp$ is a subspace of $R^D$, this subspace is called the orthogonal complement of $W$. If $W$ is a subspace of $R^D$ and $W^\perp$ is a subspace of $R^D$ then

$$\dim(W^\perp) = D - \dim(W) \quad (3)$$

At any point $x$ in the vector space $V$ we can assume that the all close neighbours to $x$ that lie on the same manifold surface belong to the same subspace W or has a small angle with its projection on W. As shown in Figure (2) we can use the well-known theory above to solve the instability problem in Isomap. As shown in the figure, if we use the distance only to choose the neighbors of $x_0$, then $x_3$ will be chosen as one of $x_0$'s neighbours. If we assume the two vectors from $x_0$ to its very two nearest neighbours are independent vectors and span the subspace $W$. Then using the bases vectors of $W$ we can find the $W^\perp$. $W^\perp$ is normal to all vectors in $W$ and if $x_3$ is not on the same surface as $x_1$ and $x_2$ then angle between $W^\perp$ and $x_3$ will be smaller than 90 degrees.

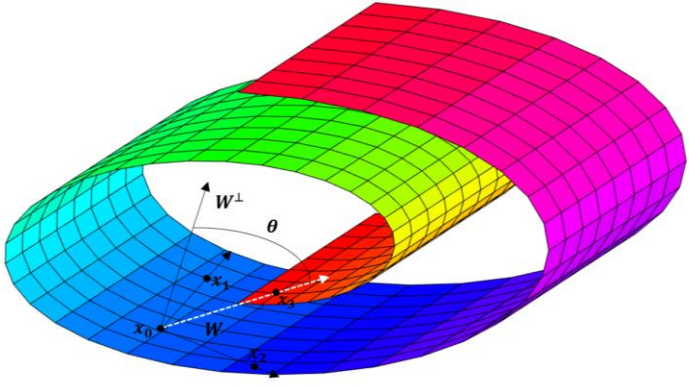

Figure (2) The white dotted vector, (bridge) $x_3$ has a small angle θ (less than 90 degrees ) with $W^\perp$ which is normal to the subspace $W$. This means that $x_3$ cannot be approximated by a vector in the subspace $W$ spanned by $x_2$ and $x_1$.

## III. THE PROPOSED ALGORITHM

Establishing the neighborhood relations is very important and if not accurate, will lead to unstable topology. The proposed algorithm is used to construct the weighted graph $G$ over the data points by establishing the neighbourhood relationship between them. The algorithm establishes the neighbourhood relationship for point $x_i$ based on two criteria;

1- The distances d($x_i, x_j$) between the pairs $x_i$ and $x_j$.
2- The angle between the vector ($x_i$-$x_j$) and any vector belong to $W^\perp$.

The other two steps are typically the same as those used in Isomap.

Table 1 shows the proposed algorithm to establish neighborhood relationships for a set of data points $M$ which are sampled from the vector space $V = R^D = \left\{x = \begin{pmatrix} x_1 \\ \vdots \\ x_D \end{pmatrix}: x_j \in R\right\}$.

As shown in Table 1 we set the dimention of the subspace $\dim(W)$ equals two. In other words, we assume that the vectors ($x_{nearest\ neighbours} - x_p$) can be well approximated as a linear combination of the bases $v_1$ and $v_2$. However, it is up to the researcher based on his undstanding of the dataset to choose the suitable $\dim(W)$. For the sake of clarification, if we choose $\dim(W) = 3$, then step 4 in the algorithm will use the three nearest neigbours to form the subspace $W$.

We should highlighted that in step 5 we choose a data point $x_q$ that is not an element of the nearest neighbours to find one of the vectors that span $W^\perp$. The intuition behind this choice is that this data point is not local to the $x_p$. We can not well approximate the vector ($x_q - x_p$) using a linear combination of the bases $v_1$ and $v_2$. Hence we can use Gram–Schmidt to to find one of the vectors that span $W^\perp$.

Table 1 the proposed algorithm

For $p = 1: M$
1. Select a data point $x_p$ ;…….
2. Determine the k nearest neighbours $\{x_{N1}, x_{N2}, \ldots, x_{Nk}\}$ which are ordered according to its distance to $x_p$.
3. Select the two nearest neighbours out of the k neighbours
   3.1 Set nearest neighbors set $NNS = \{x_{N1}, x_{N2}\}$
4. Find the two vectors $v_1$ and $v_2$ which span the subspace $W$ using Gram–Schmidt process
   4.1 Set $v_1 = (x_{N1} - x_p)$
   4.2 Set $v_2 = (x_{N2} - x_p) - \frac{\langle(x_{N2}-x_p),v_1\rangle}{\langle v_1,v_1\rangle} v_1$
5. Choose any data point $x_q$ such that $x_q \notin \{x_{N1}, x_{N2}, \ldots, x_{Nk}\}$
6. Find a vector $v_3$ which belong to $W^\perp$ using Gram–Schmidt process
   6.1 Set $v_3 = (x_q - x_p) - \frac{\langle(x_q-x_p),v_1\rangle}{\langle v_1,v_1\rangle} v_1 - \frac{\langle(x_q-x_p),v_2\rangle}{\langle v_2,v_2\rangle} v_2$
7. Find the angle between the vector $v_3$ which belong to $W^\perp$ and each of the vectors $\{x_{N3} - x_p, x_{N4} - x_p, \ldots, x_{Nk} - x_p\}$.
8. Add the data points which has an angle within the range $90° \pm \Delta\theta$ to $NNS$

End

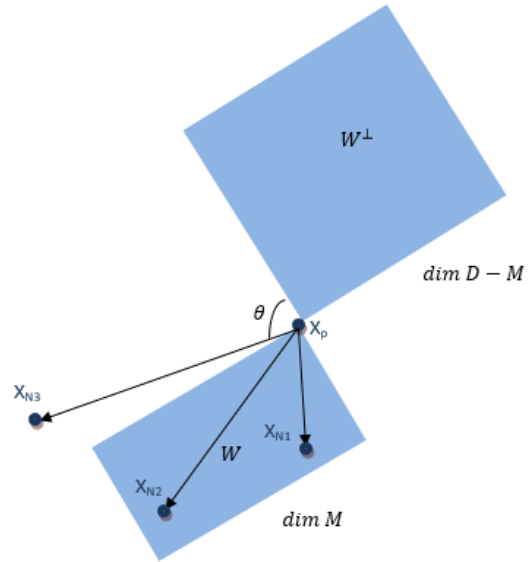

Figure (3) illustrates the $W$ subspace and the complementary subspace $W^\perp$. The $W$ subspace is spanned by $(x_{N1} - x_p)$ and $(x_{N2} - x_p)$. The vector $(x_{N3} - x_p)$ can be considered (approximated) as a vector in the subspace W if θ is close to 90 degrees

## IV. EXPERIMENTAL WORK

The experimental work consists of two parts, in the first part we used low dimensional (3D) synthesized datasets and in the second part, we used high dimensional datasets. We started with the synthesized dataset because they have well-known manifold and can be used as a ground truth.

A. *low dimensional (3D) synthesized datasets*

We have generated three synthesized datasets to test the proposed methods. These datasets are :

1- Helix.
2- Swiss Roll.
3- Punctured Sphere (the sampling is very sparse at the bottom and dense at the top).

We applied the proposed algorithm to the three datasets using different $k$. Each data point in the $k$ nearest neighbour is added to NSS, if $\theta$ within the range $90° \pm 5°$. The mapped data to lower dimensions is visually checked and it shows a good performance. The outputs of the proposed algorithm are shown in the following figures when using different $k$.

helix at different $k$. the left column is the standard Isomap and the right column is the Isomap with proposed method for choosing nearest neighbors

Figure (4) shows the low dimensional representation of the helix where the right column is the result of the proposed method. The results in the figure show that for the proposed algorithm there is no bridges between points that belong to different surfaces. Also, the results show good low dimensional representation at $k = 3$ and 10 and a perfect representation at $k = 5$.

The proposed algorithm is applied to the Swiss Roll dataset at different $k$. the $2D$ representation of the Swiss Roll is shown in Figure (5).

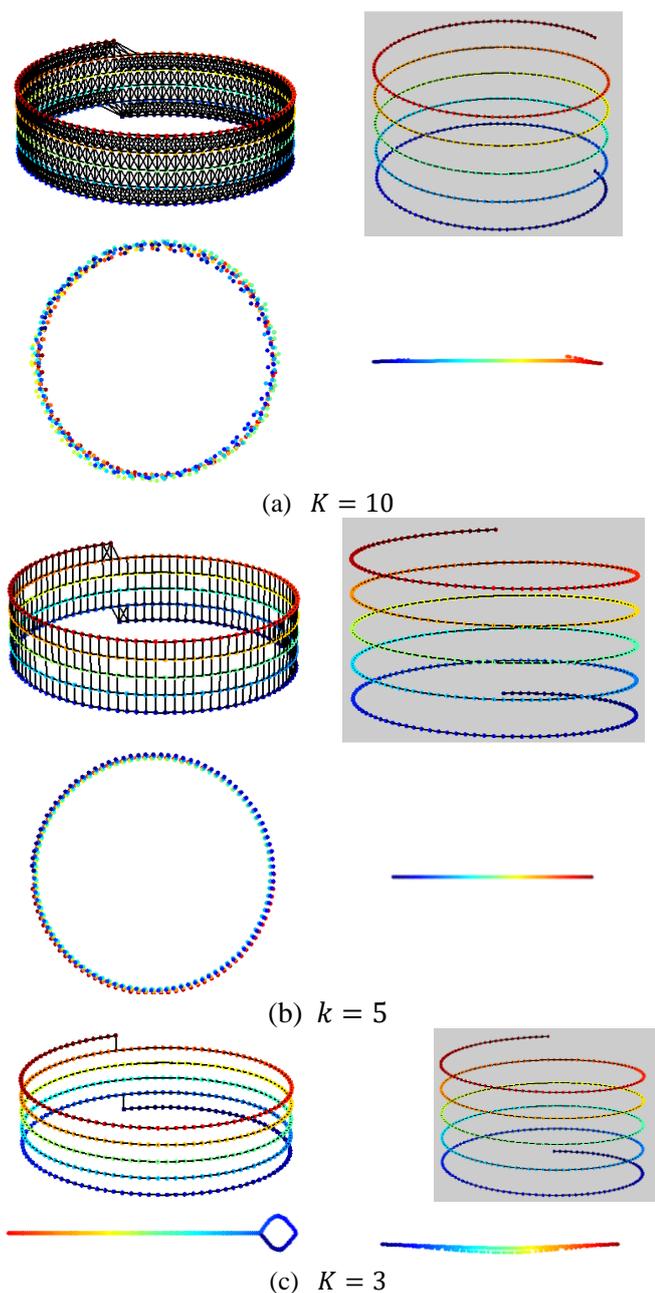

(a) $K = 10$

(b) $k = 5$

(c) $K = 3$

Figure (4) the graph and the low dimensional mapping of the

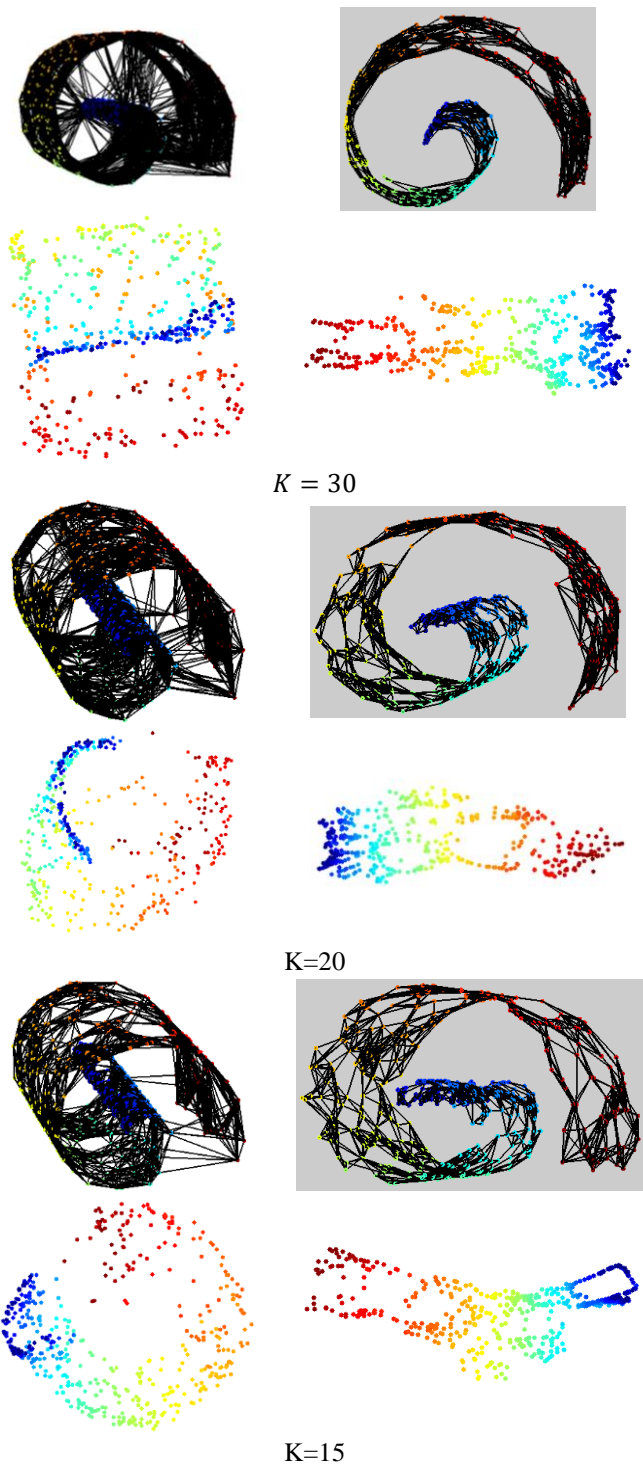

$K = 30$

K=20

K=15

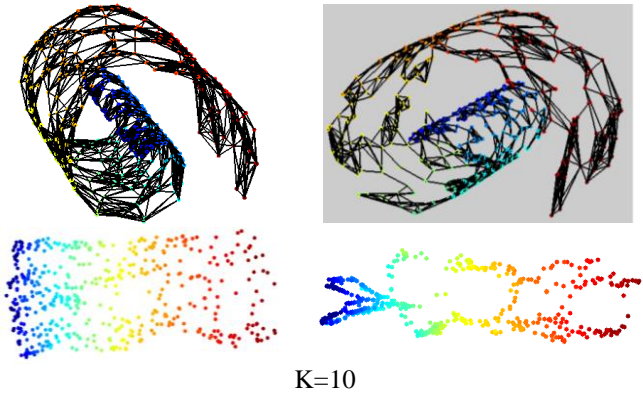

K=10

Figure (5) the 3D graph the 2D mapping of the Swiss role at different $k$. the left column is the standard Isomap and the right column is the Isomap with proposed method for choosing nearest neighbours

Figure (5) shows the 2D representation of the Swiss role where the right column is the result of the proposed method. The results show good 2D representation at $k$ =10, 15, 20 and a perfect representation at $k = 30$. The standard ISOMAP gives a perfect representation at $k = 10$ and a representation looks like linear projection at other $k$.

The proposed algorithm is applied to the punctured sphere dataset at different $K$. the 2D representation of the punctured sphere is shown in Figure (6).

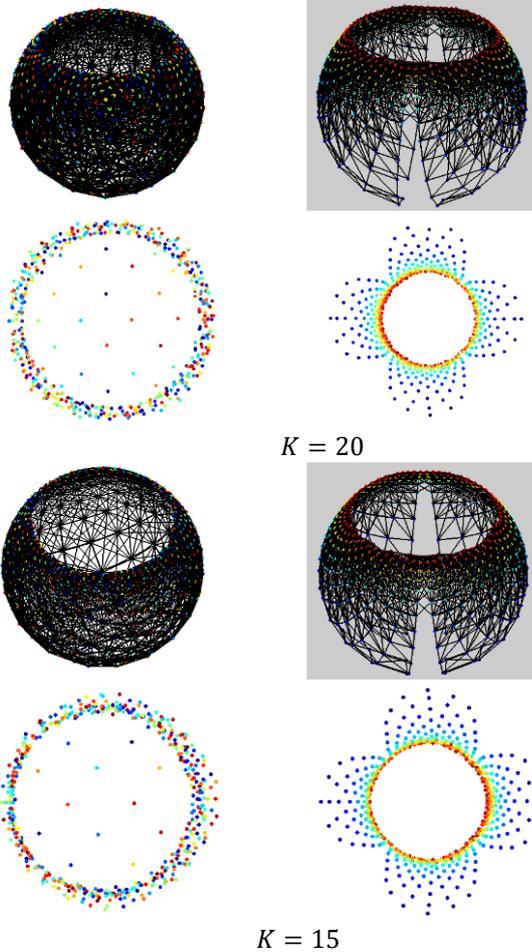

$K = 20$

$K = 15$

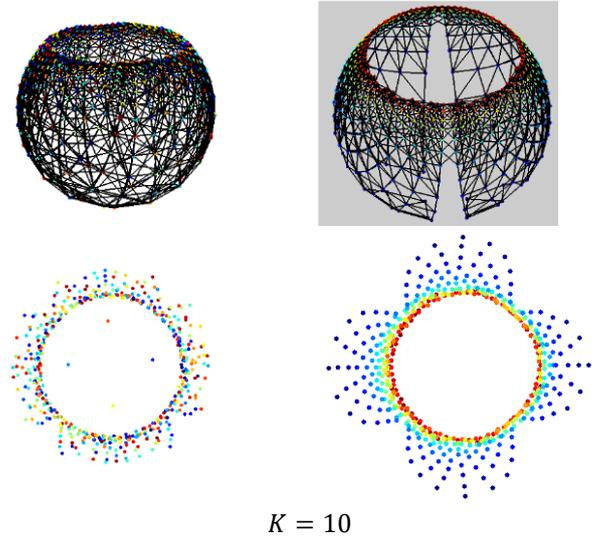

$K = 10$

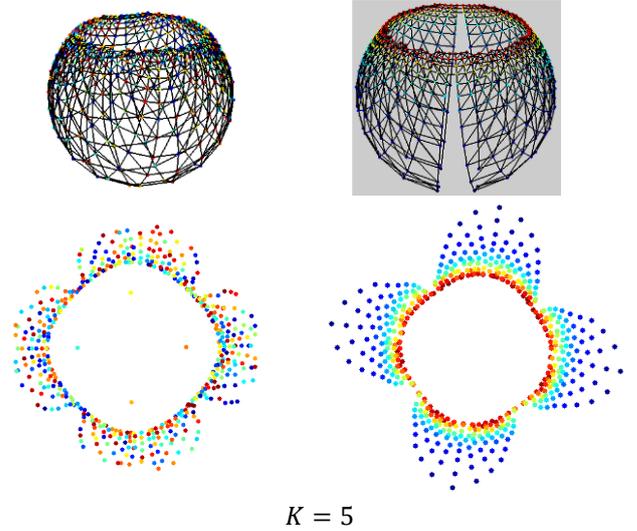

$K = 5$

Figure (6) the 3D graph and the 2D mapping of the Punctured Sphere at different $k$. The left column is the standard Isomap and the right column is the Isomap with proposed method for choosing nearest neighbors

Figure (6) shows the 2D representation of the Punctured Sphere where the right column is the result of the proposed method. The results of the proposed algorithm show good 2D representation at $k = 5$, 10, 15 and 20. The standard Isomap gives bad representation at $k = 5$, 10, 15 and 20 and a representation looks like linear projection at $k = 5$, 10, 15 and 20.

### B. High dimensional real datasets

The second part of the experiments is done using a high dimensional datasets. The datasets used are:

1- The MNIST data set consists of 60,000 grayscale images of handwritten digits. For our experiments, we selected 1,000 of the images for computational reasons [11].

2- The Olivetti faces data set consists of images of 40 individuals. The database has 10 images for each individual which were taken at different times, lighting conditions and, facial expressions (open/closed eyes, smiling / not smiling). Some of the individuals wear glasses which introduce more variation into the images.. The background of all images is homogeneous dark and the subjects in an upright, frontal position (with tolerance for some side movement) [12].

Because the datasets used in this set of experiments have high dimension and the underlying manifold of each dataset is unknown, we have used the residual variance to show the improvement of the proposed method .We compared the residual variance equation (4) of the proposed method and the standard Isomap.

$residual\ variance = 1 - R^2(D_G, D_Y)$  (4)

where $D_Y$ is the matrix of Euclidean distances in the low-dimensional embedding recovered by the algorithm, $D_G$ is the graph distance matrix and $R$ is the standard linear correlation coefficient. Figure (7) and figure (8) show the $2D$ representation of the MNIST handwritten digits at different $k$. The figures show that the proposed approach give better result where similar handwritten digits are clustered together and at the same time there is a clear separation between clusters.

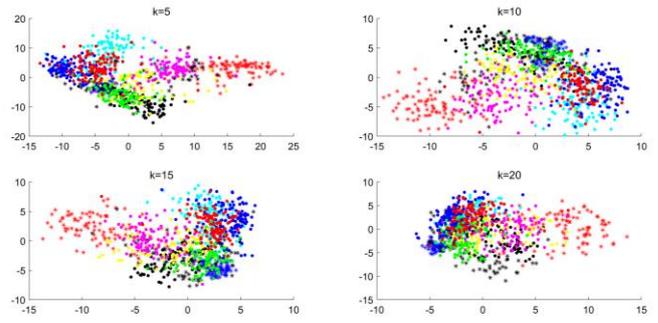

Figure (8) The standard method

In order to further show the improved performance of our proposed approach over the standard Isomap we calculated the residual variance at different $k$ for both approaches. Figure (9) shows the residual variance of the two methods. Comparing the residual variance of the two methods shows that the proposed algorithm has a better residual variance.

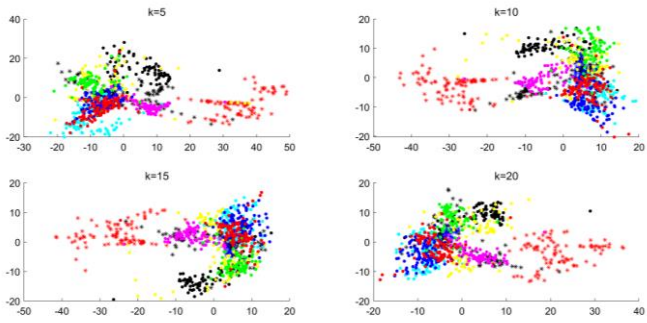

Figure (7) The proposed method

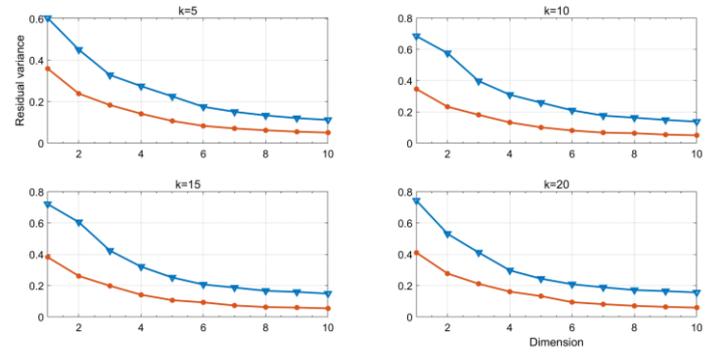

Figure (9) Comparison of the residual variance between the two method. The red curve is the proposed method and the blue curve is the standard Isomap

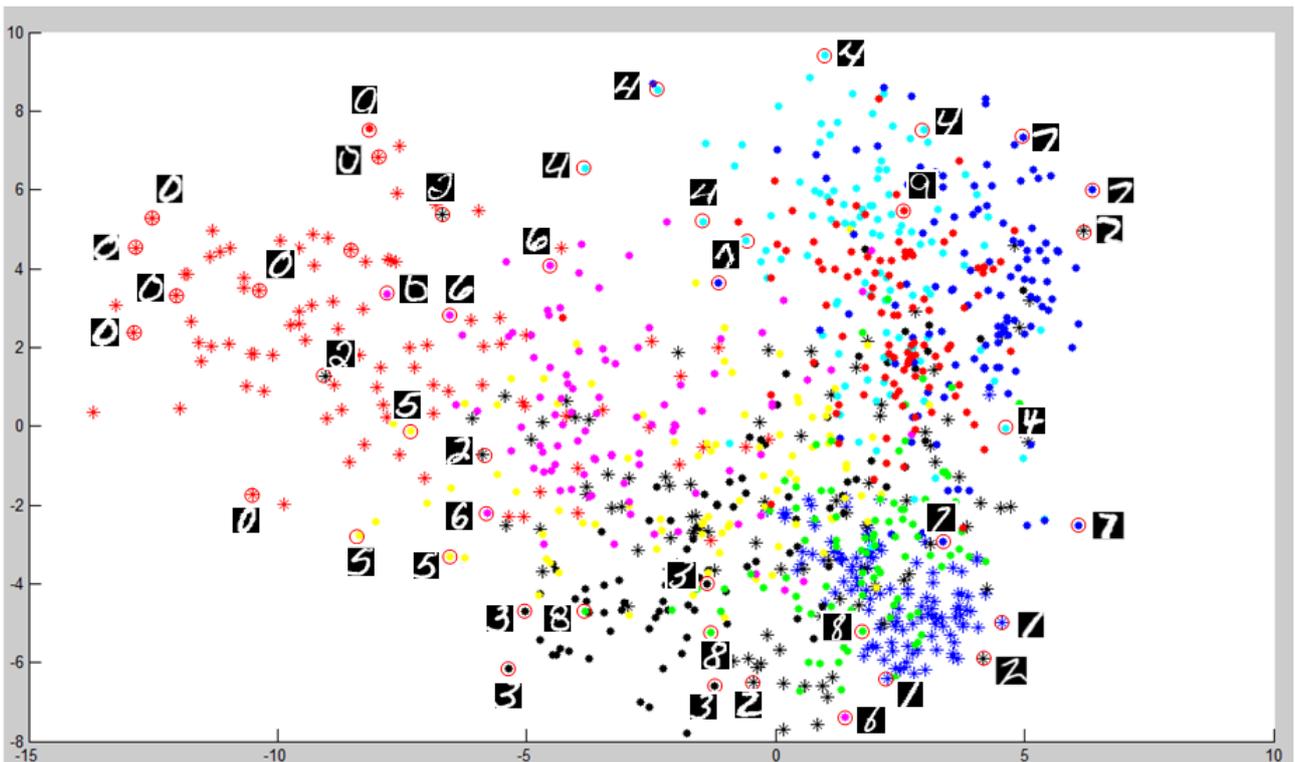

Figure (10) 2D embedding using the standard Isomap at k=15. Images of the digits mapped into the 2D embedding space. Representative digits are shown next to circled points in different parts of the space.

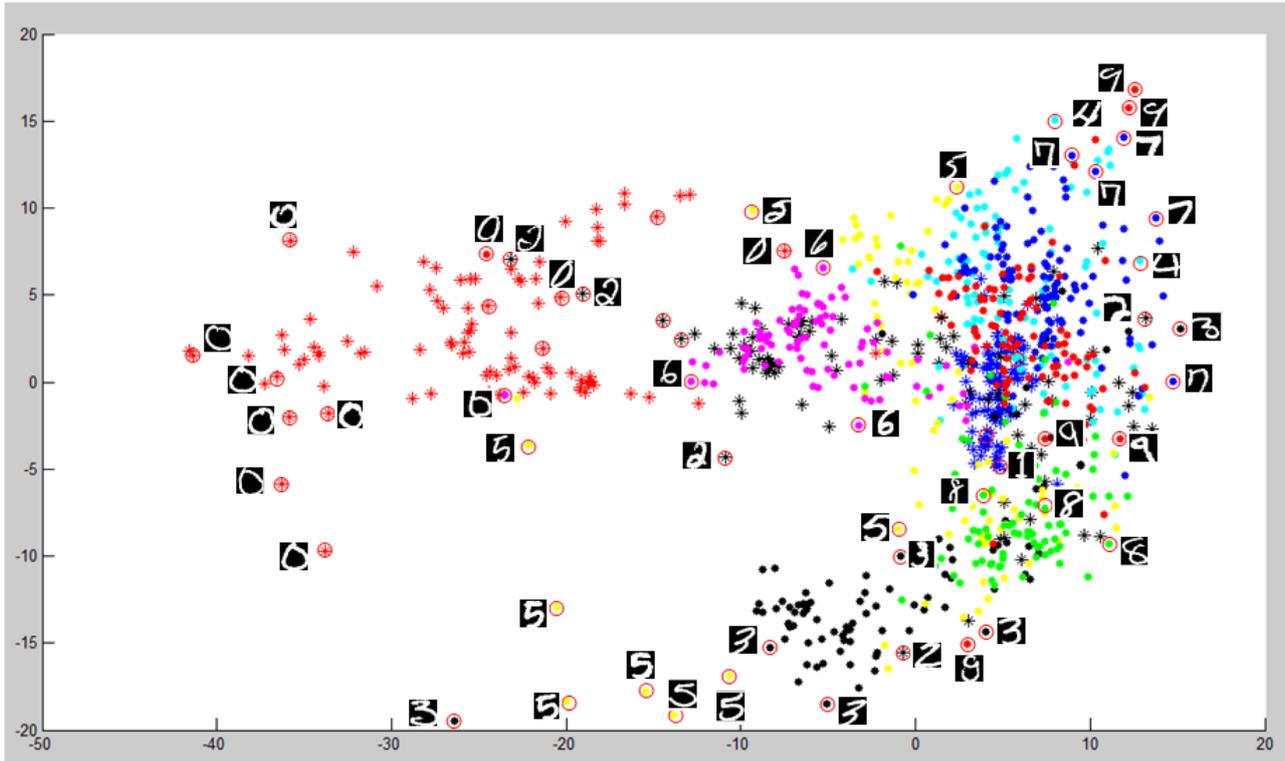

Figure (11) 2D embedding using the proposed method at k=15. Images of the digits mapped into the 2D embedding space. Representative digits are shown next to circled points in different parts of the space

We applied the both the standard Isomap and the proposed method to the Olivetti faces dataset using different $k$. the $2D$ representations of the dataset using both algorithms are shown in figures (12-13). We show representations which has the best residual variance in each algorithm. Figure (14) shows a comparison between the residual variance of the two methods when applied to the Olivetti Faces the red curve is the best residual variance for the proposed method (at $k = 13$) and the blue curve is the best residual variance for the standard Isomap. The figure shows that the proposed method has lower residual variance which means it has better performance.

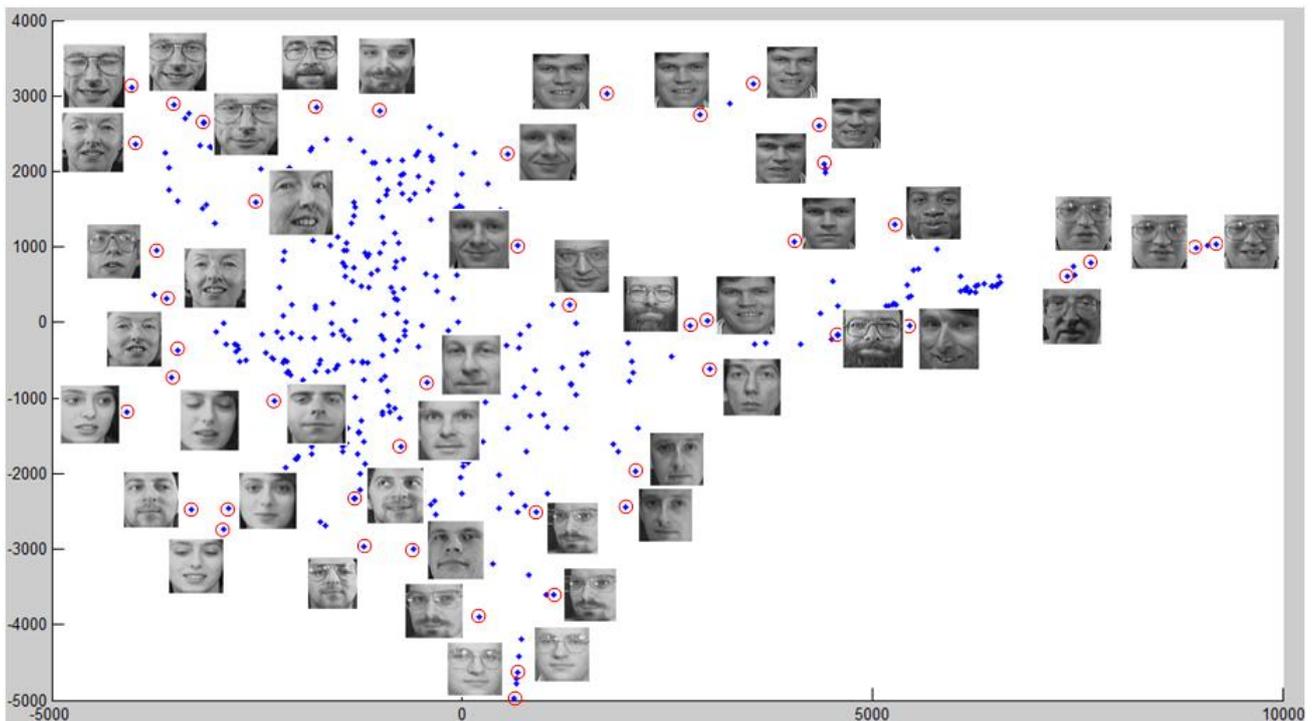

Figure (12) 2D embedding using the proposed method at k=13. Images of the faces mapped into the 2D embedding space. Representative faces are shown next to circled points in different parts of the space

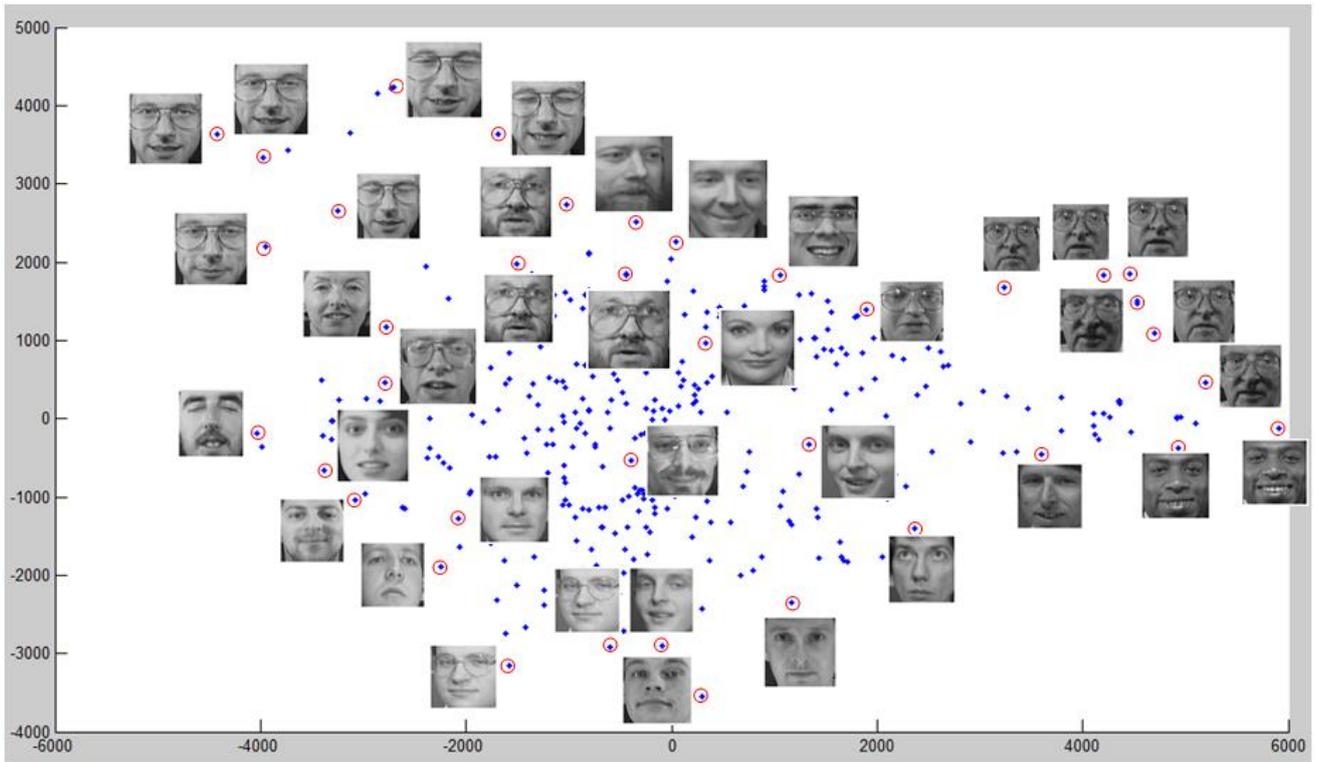

Figure (13) 2D embedding using the proposed method at k=7. Images of the faces mapped into the 2D embedding space. Representative faces are shown next to circled points in different parts of the space

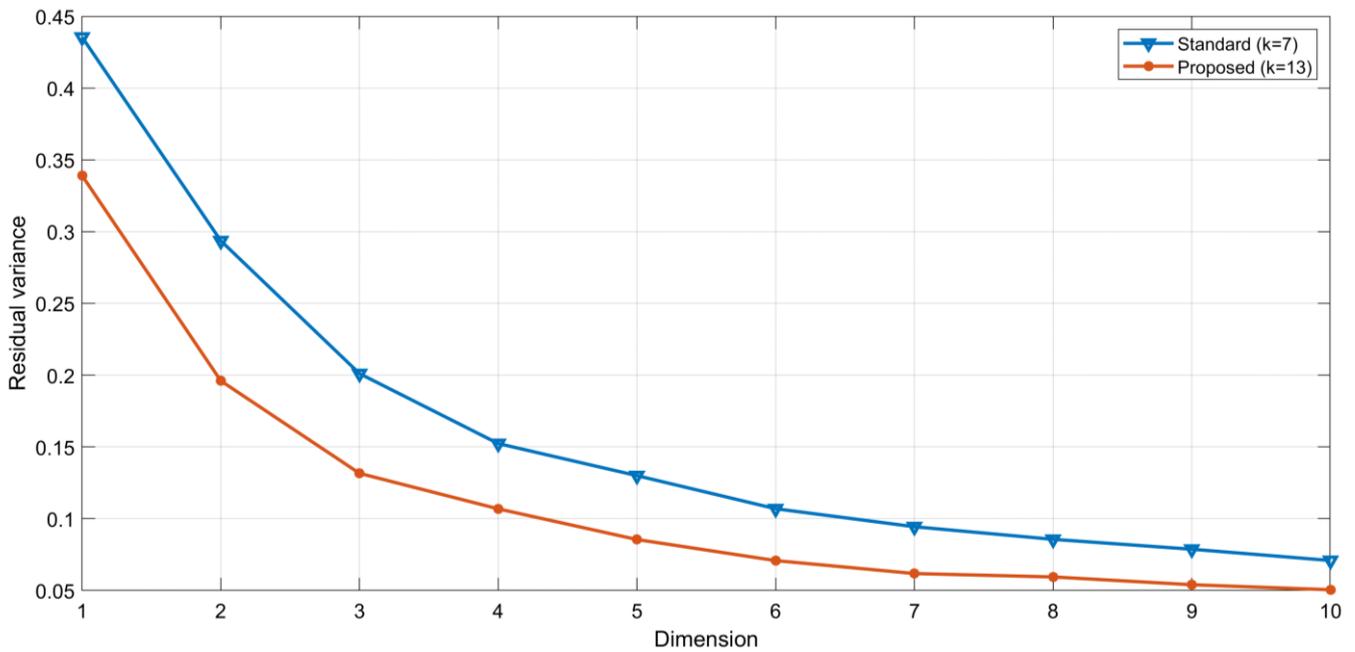

Figure (14) Comparison of the residual variance between the two methods. The red curve is the proposed method and the blue curve is the standard Isomap

## V. CONCLUSION

Dimension reduction is very important technique where it helps discovering the underling structure of the data manifold. Isomap is one of the best reduction algorithms but it suffers from topology instability. This instability arises when the manifold surfaces are close to each other at some points which introduce bridges and lead to drastically different (and incorrect) low-dimensional embedding. We proposed an algorithm based on the subspace and orthogonal complementary subspace to solve the instability problem. In our algorithm the neighbors are chosen based on both their distance and angle with the orthogonal complementary space. The proposed algorithm is tested with both synthesized datasets and high dimensional real datasets. The results show the improved stability of our approach over the standard Isomap and its lower residual variance.